\PassOptionsToPackage{table,xcdraw}{xcolor}
\PassOptionsToPackage{hyphens}{url}
\documentclass{article} %
\usepackage{preprint,times}

\usepackage{amsmath,amsfonts,bm}

\def\eqref#1{equation~\ref{#1}}

\def\1{\bm{1}}

\DeclareMathAlphabet{\mathsfit}{\encodingdefault}{\sfdefault}{m}{sl}
\SetMathAlphabet{\mathsfit}{bold}{\encodingdefault}{\sfdefault}{bx}{n}

\usepackage[T1]{fontenc}
\usepackage{hyperref}
\usepackage{url}

\usepackage{graphicx}
\usepackage{booktabs}
\usepackage{multirow}
\usepackage{array}
\usepackage{xcolor}
\usepackage{colortbl}
\usepackage{arydshln}
\usepackage{amsmath,amssymb}
\usepackage{mathtools}
\usepackage{bm}
\usepackage{enumitem}
\usepackage{microtype}
\usepackage{placeins}
\title{No Concept Escapes the Audit: Auditing-Aware Unlearning for Verifiable Concept Erasure in Diffusion Models}

\author{Kaiyuan Deng$^{1*}$\quad Yuchen Li$^{2*}$\quad Gen Li$^{3}$\quad Yang Xiao$^{4}$ \\
\textbf{Geng Yuan$^{5}$\quad Xiaoyong Yuan$^{3}$\quad Bo Hui$^{4}$\quad Xiaolong Ma$^{1\dagger}$} \\
\normalfont $^1$The University of Arizona\quad $^2$Peking University\quad $^3$Clemson University \\
\normalfont $^4$The University of Tulsa\quad $^5$University of Georgia \\
\normalfont \texttt{kaiyuan0415@arizona.edu} \\
\normalfont $^*$Equal contribution\quad $^\dagger$Corresponding author}

\ppfinalcopy
\begin{document}

\maketitle

\begin{abstract}
\label{sec:abs}
Text-to-image diffusion models can generate prohibited content, which motivates concept erasure through machine unlearning. Most erasure methods intervene at the text interface, through prompt modification or localized updates to text-conditioning weights, and they are evaluated by what the model outputs for given prompts. Such evaluation cannot see what the network still encodes. Latent-space auditing, which bypasses text conditioning and probes the denoising network directly, shows that erased concepts remain recoverable from internal representations. We find that this also holds for methods built to be robust against adversarial prompts, and that the problem grows with the number of erased concepts. We propose \textbf{\underline{A}uditing-Aware Unlearning for \underline{V}erifiable \underline{C}oncept \underline{E}rasure in Diffusion Models~(AVCE)}, a framework that grounds erasure in the model's latent representations. AVCE audits the embedding neighborhood of each concept and condenses the discovered vulnerable directions into an anchor at the weakest geometric point. It edits cross-attention and self-attention projections in closed form at this anchor, then fine-tunes the two pathways with pathway-level auditing losses, using orthogonal gradient projection to consolidate multiple concepts. Experiments on SD v1.5, SDXL, and Flux 1.0 across object, explicit-content, and artistic-style unlearning show that AVCE reduces attack success rates by $\bm{4.92\times}$ and improves auditing scores by $\bm{3.84\times}$ over the strongest baseline, while preserving competitive generation quality.
\end{abstract}

\section{Introduction}
\label{sec:intro}
Text-to-image (T2I) diffusion models generate diverse, high-fidelity images from natural language descriptions~\citep{rombach2022high, ma2024sit, podell2023sdxl, saharia2022photorealistic, nichol2021glide, ding2022cogview2, zhou2022towards}. Their wide deployment also raises ethical and legal risks, including copyright infringement, unauthorized trademark generation, and the synthesis of malicious imagery~\citep{kazerouni2022diffusion, xing2024survey, jang2022knowledge, wang2024machine, nguyen2022survey, weidinger2021ethical, brundage2018malicious, tolosana2020deepfakes, solaiman2021process}.

Machine unlearning (MU) addresses these risks by removing specific knowledge from a trained model without retraining it from scratch~\citep{bourtoule2021machine, guo2019certified, graves2021amnesiac, xu2024machine, marino2025bridge, zhang2023review,tu2025sdwv, feng2025fg}. In T2I generation, MU takes the form of concept erasure: the mapping between text prompts and a target visual concept is severed so that the model no longer renders the concept. Existing methods erase individual objects, styles, or identities~\citep{jia2023model, kumari2023ablating, chavhan2024conceptprune, gandikota2023erasing, zhang2024forget, schramowski2023safe, fan2023salun, heng2023selective}, and recent work extends erasure to multiple concepts at once~\citep{lu2024mace, lyu2024one, li2025sculpting, gandikota2024unified, gong2024reliable, deng2026forget}.

\begin{figure}[t]
  \centering
  \includegraphics[width=\linewidth]{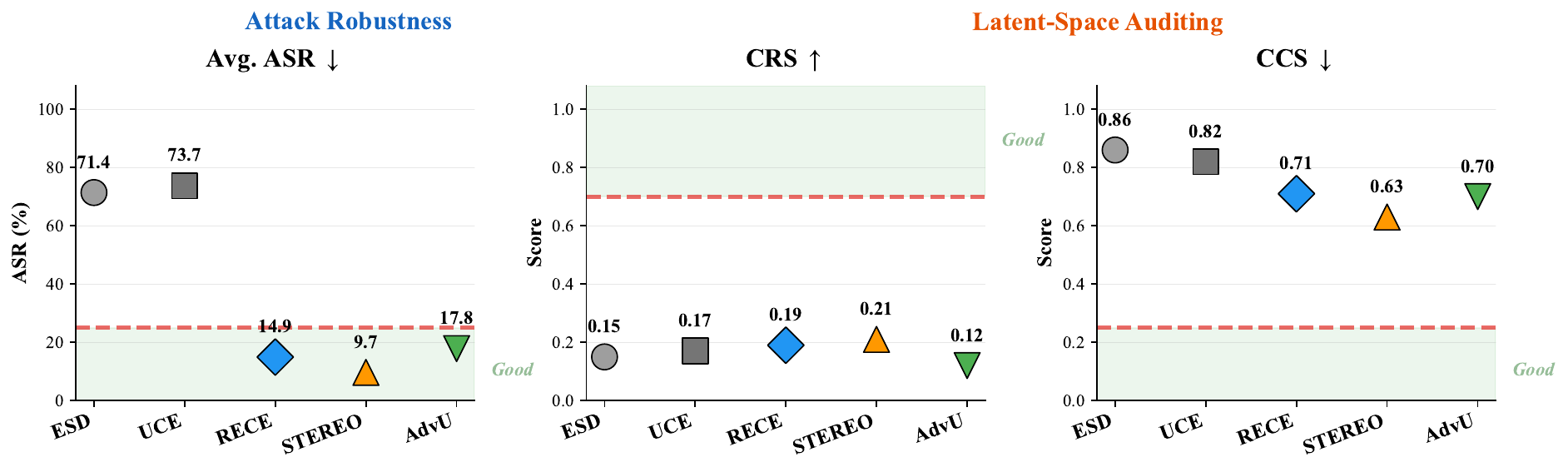}
\caption{The illusion of robustness on SD v1.5 (erasing the concept ``golf ball''). Robust methods block adversarial text inputs (left), yet latent-space auditing, which bypasses text conditioning, shows that the visual features of the target concept remain in the model (middle, right).}
  \label{fig:show}
\end{figure}

Whether an erased concept has actually been removed is usually judged from model outputs. Benchmarks~\citep{moon2025holistic,zhang2024unlearncanvas} score the images generated for a set of prompts, and adversarial evaluations~\citep{zhang2024generate,chin2023prompting4debugging,pham2023circumventing,tsai2023ring} search for prompts or embeddings that make the concept reappear. In response to these attacks, robust erasure methods harden the text-to-image mapping through adversarial training~\citep{kim2024race,zhang2024defensive}, erasure guided by inverted worst-case embeddings~\citep{li2026aegis,srivatsan2025stereo}, or closed-form re-erasure of adversarial embeddings~\citep{gong2024reliable}. These defenses bring attack success rates below those of non-robust baselines (Fig.~\ref{fig:show}, left).

Low attack success, however, does not establish that the concept is gone. The latent-space audit of \citet{sharma2024unlearning} bypasses text conditioning altogether: it injects a concept-informative latent code at an intermediate denoising step (partial diffusion) and measures, through the Concept Retrieval Score (CRS) and the Concept Confidence Score (CCS), whether the unlearned model still reconstructs the concept (Appendix~\ref{app:audit}). Under this audit, methods that block adversarial prompts still retain the concept in their internal representations (Fig.~\ref{fig:show}, middle and right; Section~\ref{sec:exp}). The cause is structural. These methods act mainly on the text encoder or on the cross-attention projections: the key and value projections of the cross-attention layers hold 2.23\% of the UNet parameters in SD v1.5 and 13.27\% in SDXL. Such edits sever the pathway from text to concept and leave intact the capacity of the denoising network to reconstruct concept-related patterns, which is the capacity that partial diffusion probes. \emph{Attack robustness therefore shows only that the evaluated attacks fail to recover the erased concept; it does not establish that concept information has been removed from the learned representations.} The discrepancy widens in multi-concept unlearning, where existing methods degrade as the number of erased concepts grows.

We therefore design erasure around the audit rather than around the attack surface, and propose \textbf{\underline{A}uditing-Aware Unlearning for \underline{V}erifiable \underline{C}oncept \underline{E}rasure in Diffusion Models~(AVCE)}, a framework that removes a concept from both the text-conditioning pathway and the visual pathway of the denoising network. AVCE proceeds in three stages. It first searches the embedding neighborhood of each concept for directions along which the concept can be revived, and condenses them into a \emph{Verifiable Vulnerability Anchor} located at the weakest point of that neighborhood. It then performs \emph{Analytical Editing} at this anchor: a closed-form rewrite of the cross-attention projections severs the association between text and concept, and a closed-form shrinkage of the self-attention projections suppresses the residual visual patterns of the concept. Finally, \emph{Audit-Guided Refinement} fine-tunes the two pathways in alternation with pathway-level auditing losses that remove the non-linear leakage left by the linear edits, and applies orthogonal gradient projection so that multiple concepts can be erased within one model.

Our contributions are as follows:
\begin{itemize}[label=$\bullet$]
    \item \textbf{The Illusion of Robustness.} We apply latent-space auditing to robust concept erasure methods and show that blocking adversarial prompts does not remove the concept: once text conditioning is bypassed, its visual features remain recoverable from the internal representations of the model.
    \vspace{0.3em}
    \item \textbf{Auditing-Aware Verifiable Concept Erasure.} We propose AVCE, which moves erasure from the text-conditioning interface to the internal representations of the model. AVCE combines a Verifiable Vulnerability Anchor obtained by auditing the embedding space, closed-form editing of both cross-attention and self-attention projections, and fine-tuning with pathway-level auditing losses.
    \vspace{0.3em}
    \item \textbf{Scalable Multi-Concept Erasure.} AVCE handles inter-concept interference with difficulty-aware sampling and orthogonal gradient projection. On SD v1.5, SDXL, and Flux 1.0, AVCE reduces attack success rates by $\bm{4.92\times}$ and improves auditing scores by $\bm{3.84\times}$ over the strongest baseline.
\end{itemize}

\section{Related Work}
\label{sec:related}
\noindent\textbf{Concept unlearning in diffusion models.}
Existing methods fall into three groups. Training-based methods, such as ESD~\citep{gandikota2023erasing}, FMN~\citep{zhang2024forget}, SA~\citep{heng2023selective}, AC~\citep{kumari2023ablating}, SalUn~\citep{fan2023salun}, and FADE~\citep{thakral2025fine}, fine-tune model weights to remove a concept through negative guidance or distribution alignment. Training-free methods such as ConceptPrune~\citep{chavhan2024conceptprune} identify and prune the neurons responsible for a concept without further optimization. Guidance-based methods such as SLD~\citep{schramowski2023safe} steer the sampling trajectory away from undesired content and leave the pretrained weights unchanged.
Multi-concept erasure follows three directions. Fine-tuning approaches, including AGE~\citep{bui2025fantastic}, CoE~\citep{li2025one}, and Sculpting Memory~\citep{li2025sculpting}, iteratively optimize model weights or masks to balance forgetting and preservation. Analytical methods such as UCE~\citep{gandikota2024unified} and ScaPre~\citep{deng2026forget} modify weight matrices in closed form. Adapter-based methods, including MACE~\citep{lu2024mace}, SPM~\citep{lyu2024one}, and GLOCE~\citep{lee2025localized}, attach lightweight modules such as LoRAs or one-dimensional adapters that suppress target concepts while the backbone remains fixed.

\vspace{0.3em}
\noindent\textbf{Robust unlearning.}
AdvUnlearn~\citep{zhang2024defensive} integrates adversarial training into concept erasure to counter prompt attacks while preserving utility. RACE~\citep{kim2024race} mitigates adversarial text embeddings with an efficient single-timestep adversarial training scheme. STEREO~\citep{srivatsan2025stereo} first identifies vulnerable embeddings and then performs anchor-guided erasure to resist embedding-space inversion. RECE~\citep{gong2024reliable} iterates closed-form solutions for rapid and robust alignment without extensive fine-tuning. AEGIS~\citep{li2026aegis} addresses the trade-off between robustness and retention in data-free settings through target-guided erasure and gradient synergy. AVCE differs from these methods in where it intervenes: it edits self-attention projections in addition to cross-attention projections, and it uses pathway-level auditing signals during fine-tuning.

\vspace{0.3em}
\noindent\textbf{Evaluating concept erasure.}
Erasure is evaluated at three levels. Output-level benchmarks~\citep{moon2025holistic,zhang2024unlearncanvas} score the images generated for fixed prompt sets with concept-specific classifiers. Attack-level evaluations search the text interface for inputs that revive the concept, through genetic search over discrete tokens~\citep{tsai2023ring}, optimization of token perturbations against the diffusion objective~\citep{zhang2024generate,chin2023prompting4debugging}, or direct inversion of the erased embedding~\citep{pham2023circumventing}. Both levels observe the model only through its text interface. Representation-level auditing instead bypasses that interface: the partial-diffusion protocol of \citet{sharma2024unlearning} injects a concept-informative latent at an intermediate denoising step and measures, through CRS and CCS, whether the unlearned network still reconstructs the concept. We adopt this protocol as our primary evaluation and, unlike prior work, also use its signal inside the erasure procedure.

\section{Method}
\label{sec:method}
\begin{figure}[t]
  \centering
  \includegraphics[width=\linewidth]{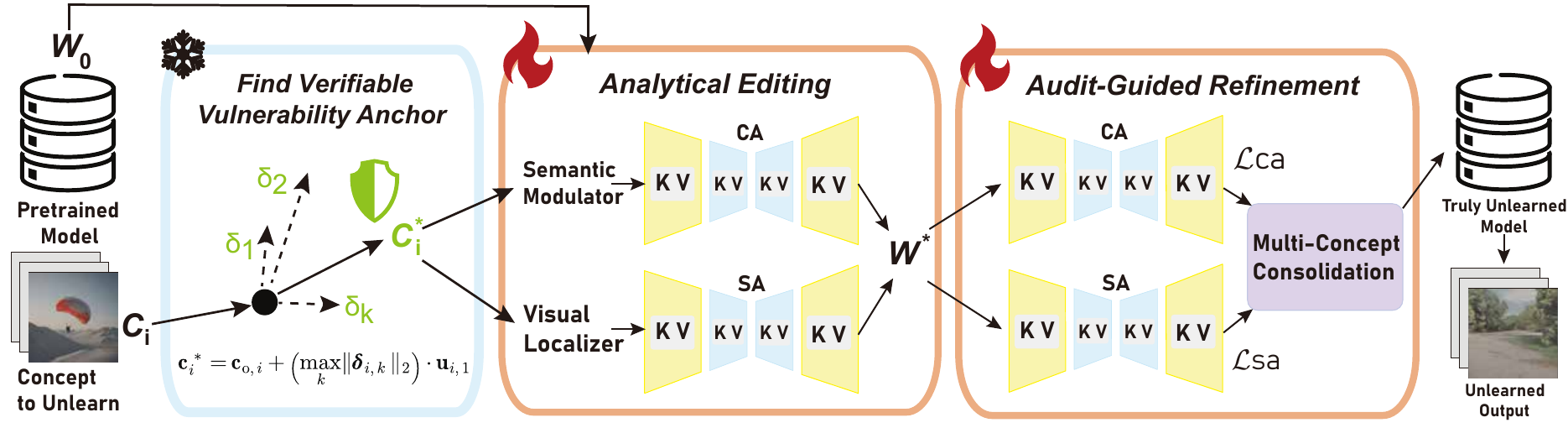}
\caption{Overview of the AVCE framework. The Verifiable Vulnerability Anchor is obtained by auditing the embedding space for its weakest geometric point. Analytical Editing rewrites cross-attention via a Semantic Modulator and shrinks self-attention via a Visual Localizer. Audit-Guided Refinement closes residual leakage through pathway-level fine-tuning with multi-concept gradient consolidation.}
  \label{fig:method}
\end{figure}

Fig.~\ref{fig:method} gives an overview of AVCE. The framework has three stages. We first audit the embedding neighborhood of each target concept and construct a Verifiable Vulnerability Anchor (Section~\ref{sec:anchor}). We then edit the cross-attention and self-attention projections in closed form at this anchor (Section~\ref{sec:editing}). Finally, we fine-tune the edited model with pathway-level auditing losses and consolidate multiple concepts through orthogonal gradient projection (Section~\ref{sec:ft}).

\subsection{Verifiable Vulnerability Anchor}
\label{sec:anchor}

Existing unlearning methods apply erasure at the original prompt embedding and leave robustness to separate, post-hoc defenses. The neighborhood of that embedding, however, may contain directions along which the erased concept can be revived. We therefore audit this neighborhood before editing and replace the original embedding with an anchor that summarizes the vulnerable directions found.

For a set of $N$ target concepts $\{\mathcal{C}_i\}_{i=1}^N$, let $c_{\mathrm{o},i} \in \mathbb{R}^{d_{\mathrm{in}}}$ be the original text embedding of the $i$-th concept. We optimize probe embeddings $c_{\mathrm{a},i}$ that reproduce the generative behavior of $\mathcal{C}_i$ while lying far from $c_{\mathrm{o},i}$. A search through the full denoising network alone tends to return embeddings that revive the concept by re-routing the attention maps of the cross-attention (CA) layers, that is, through prompt alignment, rather than through the visual content that the CA layers inject and the self-attention (SA) layers propagate. The auditing objective therefore also evaluates each candidate with the CA attention maps held fixed.
To cover diverse vulnerable directions, we run $K$ randomly initialized searches. The objective and the gradient update for the $k$-th search at iteration $j$ are
\begin{align}
\label{eq:audit_loss}
\mathcal{L}_{\mathrm{audit}}(c) \;=\;& \mathbb{E}_{z_t, t} \Big[ \|\epsilon_{\theta, t}(z_t, c) - \epsilon_{\theta, t}(z_t, c_{\mathrm{o},i})\|_2^2 \Big] - \tau \|c - c_{\mathrm{o},i}\|_2 \notag\\
&+\; \mu \, \mathbb{E}_{z_t, t} \Big[ \|\tilde{\epsilon}_{\theta, t}(z_t, c) - \tilde{\epsilon}_{\theta, t}(z_t, c_{\mathrm{o},i})\|_2^2 \Big],
\end{align}
\begin{equation}
\label{eq:audit_update}
c_{\mathrm{a},i,k}^{(j+1)} \;=\; c_{\mathrm{a},i,k}^{(j)} - \beta \, \nabla_{c} \mathcal{L}_{\mathrm{audit}}\big(c_{\mathrm{a},i,k}^{(j)}\big),
\end{equation}
where $z_t$ is the noisy latent at timestep $t$, $\epsilon_{\theta, t}$ is the denoising network, $\tilde{\epsilon}_{\theta, t}$ denotes its variant in which every CA layer computes its attention maps from $c_{\mathrm{o},i}$ (keys from $c_{\mathrm{o},i}$) and its values from the input condition $c$, so that $\tilde{\epsilon}_{\theta, t}(z_t, c_{\mathrm{o},i}) = \epsilon_{\theta, t}(z_t, c_{\mathrm{o},i})$, $\tau > 0$ is the repulsion coefficient, $\mu \ge 0$ controls the auditing strength, and $\beta$ is the step size. The first term favors embeddings whose full-network prediction matches that of the original concept. The second term pushes the search away from $c_{\mathrm{o},i}$ so that it reaches distant regions of the embedding space. The third term holds the attention maps at those of $c_{\mathrm{o},i}$, so a candidate can reduce it only through the content it injects; it thereby penalizes candidates that revive the concept only through prompt alignment.

A search counts as successful when its probe reproduces the noise prediction of the original concept substantially better than the null prompt does, which excludes probes that drifted into unrelated regions of the embedding space (Appendix~\ref{app:AVCE_training}). From the successful searches $\{c_{\mathrm{a},i,k}\}_{k=1}^K$, we compute the perturbations $\delta_{i,k} = c_{\mathrm{a},i,k} - c_{\mathrm{o},i}$ and stack them into a residual matrix $\Delta_i$. The \emph{Verifiable Vulnerability Anchor} $c_i^\star$ places the largest perturbation magnitude along the principal direction $u_{i,1}$, the leading left singular vector of $\Delta_i$, whose sign is fixed to agree with the mean perturbation:
\begin{equation}
\label{eq:anchor_def}
c_i^\star \;=\; c_{\mathrm{o},i} \;+\; \big(\max_k \|\delta_{i,k}\|_2\big) \, u_{i,1}.
\end{equation}
The anchor condenses the audited directions into a single point. All subsequent editing is applied at $c_i^\star$ instead of $c_{\mathrm{o},i}$, that is, at the weakest point found by the audit.

\subsection{Analytical Editing for a Safer Model Space}
\label{sec:editing}

Given the anchor $c_i^\star$, we modify the model through two closed-form edits that act on disjoint parameter sets. The first rewrites the cross-attention projections to remove the association between text and concept. The second shrinks the self-attention projections to suppress the residual visual patterns of the concept. Both edits are computed in one shot and provide the initialization for the subsequent fine-tuning.

\subsubsection{(I) Cross-Attention Analytical Erasure}
\label{sec:ca_editing}
Cross-attention layers map text embeddings to visual features and are the main channel through which concept semantics enter generation. We rewrite their projections in closed form so that the concept association is removed while unrelated generative capabilities are preserved. Let $W^\star_{\mathrm{ca}} \in \mathbb{R}^{d_{\mathrm{out}}\times d_{\mathrm{in}}}$ denote the edited weight for erasing $N$ concepts. The erasure target is the unconditional response $v = W_0 c_{\emptyset}$, where $W_0$ is the original projection and $c_{\emptyset}$ is the end-of-text embedding of the null prompt; every erased concept is mapped to this same target.

\vspace{0.3em}
\noindent\textbf{Semantic Modulator.}
To limit collateral damage, we select output channels according to statistical stability and information-theoretic discriminability. For a calibration set of $P$ target-concept samples, let $a_{i,r}^{(p)}$ denote the activation of channel $r$ on the $p$-th sample, with mean $\bar{a}_{i,r}$ and variance $\sigma_{i,r}^2$. We compute a Fisher-style stability score $f_{i,r} = \bar{a}_{i,r}^2 / (\sigma_{i,r}^2 + \varepsilon)$ and the Symmetric Uncertainty $m_{i,r} = 2\,I(q_{i,r}; y_i) / (H(q_{i,r}) + H(y_i) + \varepsilon)$, where $q_{i,r} = \mathbf{1}[a_{i,r} > \mathbb{E}[a_{i,r}]]$ is a binary activation state, $y_i \in \{0,1\}$ is the target-vs-neutral label, $I(\cdot;\cdot)$ is mutual information, and $H(\cdot)$ is Shannon entropy. Temperature-scaled sigmoids map the two scores to bounded multipliers $\rho_{i,r}, \alpha_{i,r} \in [0,1]$. Their product $g_{i,r} = \alpha_{i,r} \cdot \rho_{i,r}$ forms a sparse vector $g_i$ that confines weight updates to robust semantic pathways.

\vspace{0.3em}
\noindent\textbf{Analytical Erasure.}
We combine the output-side Semantic Modulator with an input-side constraint built from the perturbations collected during anchor construction. Stacking the perturbations of all successful searches, $M \le NK$ in total, into $\Delta \in \mathbb{R}^{d_{\mathrm{in}}\times M}$, the empirical covariance $H_{\mathrm{a}} = \frac{1}{M}\Delta\Delta^\top \in \mathbb{R}^{d_{\mathrm{in}}\times d_{\mathrm{in}}}$ captures the principal vulnerable directions in embedding space. To keep the response to unrelated text unchanged, we also include a preservation set $\mathcal{R}$ of token embeddings $c_r$ taken from neutral prompts, following UCE~\citep{gandikota2024unified}. We define the input geometry operator $Q_2$ and the modulated target matrix $Q_1$ as
\begin{gather}
\label{eq:Q2_def}
Q_2 \;=\; \lambda I \;+\; \sum_{i=1}^{N} c_i^\star {c_i^\star}^\top \;+\; 2 H_{\mathrm{a}} \;+\; \sum_{r \in \mathcal{R}} c_r c_r^\top \;\in\; \mathbb{R}^{d_{\mathrm{in}}\times d_{\mathrm{in}}}, \\
\label{eq:Q1_def}
Q_1 \;=\; \lambda W_0 \;+\; \sum_{i=1}^{N} \mathrm{Diag}(g_i)\, v\, {c_i^\star}^\top \;+\; \sum_{r \in \mathcal{R}} W_0\, c_r c_r^\top \;\in\; \mathbb{R}^{d_{\mathrm{out}}\times d_{\mathrm{in}}},
\end{gather}
where $\lambda > 0$ is the ridge coefficient. The edited weight is
\begin{equation}
\label{eq:closed_form_main}
W^\star_{\mathrm{ca}} \;=\; Q_1 Q_2^{-1}.
\end{equation}
Because $H_{\mathrm{a}}$ inflates the eigenvalues of $Q_2$ along vulnerable axes, $Q_2^{-1}$ attenuates the response of $W^\star_{\mathrm{ca}}$ to inputs along those directions, which limits how far an embedding in the audited neighborhood can reactivate the concept. The preservation terms keep $W^\star_{\mathrm{ca}} c_r$ close to $W_0 c_r$: text embeddings share a large common component, and without these terms the attenuation along $c_i^\star$ and the vulnerable directions also suppresses the response to unrelated prompts.

\subsubsection{(II) Self-Attention Analytical Shrinkage}
\label{sec:sa_editing}
Self-attention layers propagate and consolidate visual features across spatial positions and receive no direct text input. After the cross-attention edit, concept-specific spatial patterns remain in these layers and can still be exploited. We suppress them with a closed-form shrinkage: the patterns are first localized with the help of cross-attention, and the corresponding directions in the SA weight space are then shrunk.

\vspace{0.3em}
\noindent\textbf{Visual Localizer.}
We use cross-attention as a semantic probe to locate concept-specific SA features. For a fixed noisy latent $z_t$ at an intermediate timestep, we run two forward passes, one conditioned on concept $\mathcal{C}_i$ and one unconditional. At each SA--CA layer pair $\ell$, we record the SA input features $X_{\ell}^{(\mathrm{cond})}, X_{\ell}^{(\mathrm{uncond})} \in \mathbb{R}^{L \times d}$ and the CA output features $Y_{\ell}^{(\mathrm{cond})}, Y_{\ell}^{(\mathrm{uncond})} \in \mathbb{R}^{L \times d_{\mathrm{out}}}$, where $L$ is the number of spatial tokens. The difference between the two CA outputs yields a normalized spatial mask $w_{\ell,i}^{(s)}$, and the visual anchor $\hat{v}_{\ell,i}$ is obtained by pooling the conditional SA input with this mask:
\begin{gather}
\label{eq:ca_mask}
w_{\ell,i}^{(s)} \;=\; \frac{\|Y_{\ell,s}^{(\mathrm{cond})} - Y_{\ell,s}^{(\mathrm{uncond})}\|_2}{\sum_{s'=1}^{L} \|Y_{\ell,s'}^{(\mathrm{cond})} - Y_{\ell,s'}^{(\mathrm{uncond})}\|_2 + \varepsilon}, \quad s = 1,\dots,L, \\
\label{eq:sa_anchor}
\hat{v}_{\ell,i} \;=\; \sum_{s=1}^{L} w_{\ell,i}^{(s)} \, X_{\ell,s}^{(\mathrm{cond})} \;\in\; \mathbb{R}^{d}.
\end{gather}
The mask concentrates $\hat{v}_{\ell,i}$ on the spatial positions where the text condition changes the CA output, that is, where the concept is rendered, and down-weights the background.

\vspace{0.3em}
\noindent\textbf{Analytical Shrinkage.}
We stack the visual anchors into $\hat{V}_\ell = [\hat{v}_{\ell,1}, \dots, \hat{v}_{\ell,N}] \in \mathbb{R}^{d \times N}$ and compute the empirical visual covariance $\Sigma_{\mathrm{sa},\ell} = \frac{1}{N} \hat{V}_\ell \hat{V}_\ell^\top$. The key and value projections are then shrunk from the right in closed form:
\begin{equation}
\label{eq:sa_deflation}
W^\star_{\mathrm{sa}} \;=\; W_{\mathrm{sa}} \big(I + \kappa \, \Sigma_{\mathrm{sa},\ell}\big)^{-1},
\end{equation}
where $\kappa \ge 0$ controls the shrinkage strength. The edit compresses representations along the concept-related visual axes and leaves orthogonal directions unchanged, at the cost of one $d \times d$ matrix inverse per layer.

\subsection{Audit-Guided Refinement}
\label{sec:ft}

The analytical edits are linear, whereas the denoising network is deeply non-linear, so concept traces may persist in the interactions between layers. We therefore add a refinement phase that alternates between the two pathways and includes an auditing term in every update, so that concept removal is checked at the pathway level rather than at the output level.

Let $\epsilon_{s,t}(c)$ and $\epsilon_{b,t}(c)$ denote the noise predictions of the student (edited) model and of the frozen base teacher at timestep $t$ under text condition $c$. Let $\epsilon_{s,t}^{\setminus \mathrm{ca}}(c)$ denote the student's prediction with its cross-attention projections restored to the base weights. Since the student departs from the base model only in the edited attention projections, $\epsilon_{s,t}^{\setminus \mathrm{ca}}$ departs from the base prediction only through the self-attention pathway. For a target concept $\mathcal{C}_i$, we write $c_{\mathrm{o}}$ and $c^\star$ for its original embedding $c_{\mathrm{o},i}$ and its anchor $c_i^\star$, dropping the index $i$, and $\emptyset$ for the null condition.

Each iteration alternates between two pathway-specific updates. The first updates only the cross-attention parameters:
\begin{equation}
\label{eq:ca_ft_loss}
\mathcal{L}_{\mathrm{ca}} \;=\; \mathbb{E}_{z_t,t} \Big[ \|\epsilon_{s,t}(c_{\mathrm{o}}) - \epsilon_{b,t}(\emptyset)\|_2^2 \Big] \;+\; \lambda_{\mathrm{ca}} \, \mathbb{E}_{z_t,t} \Big[ \|\epsilon_{s,t}(c_{\mathrm{o}}) - \epsilon_{s,t}^{\setminus \mathrm{ca}}(c_{\mathrm{o}})\|_2^2 \Big].
\end{equation}
The first term moves the student's output for the concept prompt toward the unconditional output of the base model. The second term audits whether the erasure originates from the cross-attention pathway. It measures the discrepancy between the full-model output and the output with cross-attention frozen, and thereby drives the contribution of cross-attention to concept generation toward zero instead of allowing self-attention to compensate. The second update modifies only the self-attention parameters:
\begin{equation}
\label{eq:sa_ft_loss}
\mathcal{L}_{\mathrm{sa}} \;=\; \mathbb{E}_{z_t,t} \Big[ \|\epsilon_{s,t}^{\setminus \mathrm{ca}}(c_{\mathrm{o}}) - \epsilon_{s,t}^{\setminus \mathrm{ca}}(\emptyset)\|_2^2 \Big] \;+\; \lambda_{\mathrm{sa}} \, \mathbb{E}_{z_t,t} \Big[ \|\epsilon_{s,t}^{\setminus \mathrm{ca}}(c^\star) - \epsilon_{s,t}^{\setminus \mathrm{ca}}(\emptyset)\|_2^2 \Big].
\end{equation}
Both terms are computed with cross-attention frozen, so their gradients are attributable to the self-attention pathway alone. The first term moves the internal visual response of the model to $c_{\mathrm{o}}$ toward the unconditional response. The second term applies the same constraint at the anchor $c^\star$ (Section~\ref{sec:anchor}), which addresses the non-linear leakage along the direction that the closed-form edits are least likely to have covered.

\vspace{0.3em}
\noindent\textbf{Retaining unrelated generation.}
The erasure targets in Eqs.~(\ref{eq:ca_ft_loss}) and~(\ref{eq:sa_ft_loss}) do not constrain the response to unrelated text, and fine-tuning against them alone degrades generation for other prompts. Each update therefore adds a retain term on neutral prompts $c_{\mathrm{n}}$, drawn from the prompts that define the preservation set $\mathcal{R}$ (Section~\ref{sec:ca_editing}):
\begin{equation}
\label{eq:retain_loss}
\mathcal{L}_{\mathrm{ca}} \mathrel{+}= \lambda_{\mathrm{r}} \, \mathbb{E}_{z_t,t} \Big[ \|\epsilon_{s,t}(c_{\mathrm{n}}) - \epsilon_{b,t}(c_{\mathrm{n}})\|_2^2 \Big], \qquad
\mathcal{L}_{\mathrm{sa}} \mathrel{+}= \lambda_{\mathrm{r}} \, \mathbb{E}_{z_t,t} \Big[ \|\epsilon_{s,t}^{\setminus \mathrm{ca}}(c_{\mathrm{n}}) - \epsilon_{b,t}(c_{\mathrm{n}})\|_2^2 \Big].
\end{equation}

\vspace{0.5em}
\noindent\textbf{Scalable Multi-Concept Consolidation.}
When $N$ concepts are erased simultaneously, interference between their gradients can destabilize training. For each concept $i$, we maintain an exponential moving average $\tilde{\mathcal{L}}_i$ of its $\mathcal{L}_{\mathrm{sa}}$ as a difficulty score, and we sample a mini-batch of $B$ concepts at each iteration with probability
\begin{equation}
\label{eq:hcm_prob}
p_i \;=\; 0.8 \, \frac{\tilde{\mathcal{L}}_i}{\textstyle\sum_{l=1}^{N}\tilde{\mathcal{L}}_l} \;+\; \frac{0.2}{N},
\end{equation}
which prioritizes harder concepts while keeping a uniform exploration floor. Let $h_i$ denote the gradient for sampled concept $i$. For each conflicting pair ($h_i^\top h_j < 0$), we project $h_i$ onto the normal plane of $h_j$, following gradient surgery~\citep{yu2020gradient}:
\begin{equation}
\label{eq:grad_proj}
h_i^\star \;=\; h_i \;-\; \frac{h_i^\top h_j}{\|h_j\|_2^2} \, h_j.
\end{equation}
The projection removes the component of $h_i$ that opposes $h_j$, so an update for one concept does not undo progress on another.

\section{Experiment}
\label{sec:exp}
We evaluate AVCE on three concept unlearning tasks: multi-concept object unlearning (Section~\ref{exp:obj}), explicit-content unlearning (Section~\ref{exp:nude}), and artistic-style unlearning (Section~\ref{exp:art}), all under the protocol of Section~\ref{exp:setting}. Appendix~\ref{app:more} reports results on an additional backbone~(Tables~\ref{tab:obj_10_flux_app} and~\ref{tab:nsfw_flux_app}), large-scale unlearning~(Fig.~\ref{fig:scale}), and runtime~(Table~\ref{tab:time_efficiency}).

\subsection{Experimental Settings}
\label{exp:setting}
All experiments are primarily conducted on Stable Diffusion XL (SDXL)~\citep{podell2023sdxl}, with additional evaluations on Stable Diffusion v1.5 (SD v1.5)~\citep{rombach2022high} and Flux 1.0~\citep{flux2024}.
We compare AVCE against eight representative baselines:
ESD~\citep{gandikota2023erasing}, 
SPM~\citep{lyu2024one}, 
MACE~\citep{lu2024mace}, 
SM~\citep{li2025sculpting}, 
UCE~\citep{gandikota2024unified}, 
RECE~\citep{gong2024reliable}, 
STEREO~\citep{srivatsan2025stereo}, 
and AdvUnlearn (AdvU)~\citep{zhang2024defensive}.
Multi-object unlearning is evaluated on Imagenette~\citep{imagenette}, a ten-class ImageNet subset.
Explicit-content unlearning uses the I2P dataset~\citep{schramowski2023safe}.
For artistic-style unlearning, we target five styles and follow prior settings~\citep{chavhan2024conceptprune,deng2026forget}.
To evaluate attack success rate (ASR), we use concept-specific classifiers to detect whether erased concepts reappear in generated images: an ImageNet-pretrained ResNet50 for objects~\citep{chavhan2024conceptprune} and NudeNet for explicit content~\citep{bedapudi2019nudenet}.
We further assess robustness using Ring-A-Bell~\citep{tsai2023ring}~(ASR1), UnlearnDiffAtk~\citep{zhang2024generate}~(ASR2), and P4D~\citep{chin2023prompting4debugging}~(ASR3).
We also adopt Concept Confidence Score (CCS) and Concept Retrieval Score (CRS)~\citep{sharma2024unlearning} as auditing metrics to distinguish genuine erasure from latent-space concealment.
We evaluate FID and CLIP score on COCO-30K, a 30K-caption subset of MS-COCO~\citep{lin2014microsoft}, to measure generative fidelity and text-image alignment after unlearning.
AVCE is trained and evaluated on NVIDIA H100 (80\,GB) GPUs. For reproducibility, all baseline results are obtained from official open-source implementations (Appendix~\ref{app:code}).

\subsection{Object Unlearning}
\label{exp:obj}
\begin{figure}[t]
  \centering
  \includegraphics[width=\linewidth]{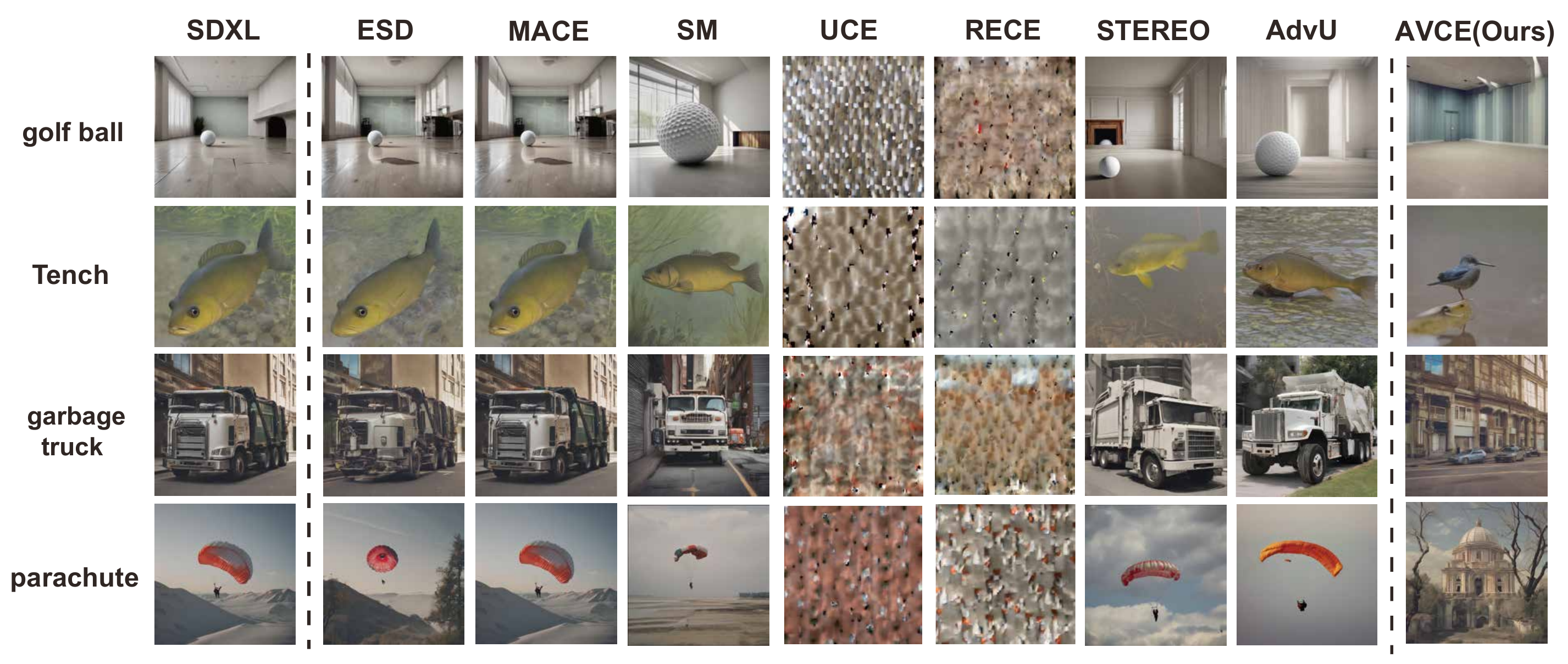}
\caption{Visual comparison of different methods for multi-concept unlearning on SDXL.}
  \label{fig:exp_img10}
\end{figure}

Erasure effectiveness is measured by Unlearn Acc, the accuracy of a pretrained ResNet50 classifier in detecting the target objects in generated images. Table~\ref{tab:obj_10_sdxl_exp} shows that AVCE attains the lowest Unlearn Acc among the methods that remain usable, with CLIP and FID on COCO-30K comparable to those of the baselines. UCE and RECE also reach zero Unlearn Acc, but only because the edited model collapses (Fig.~\ref{fig:exp_img10}), which makes them unusable in this setting.
Under the three attacks, AVCE reduces ASR by over 50 percentage points on average relative to the strongest baseline, STEREO. Fig.~\ref{fig:exp_img10} shows that most baselines still generate recognizable target objects under adversarial prompts, whereas the outputs of AVCE no longer contain them. The auditing metrics separate the methods further. Every baseline has low CRS and high CCS, which indicates that the erased concepts remain in the latent space, while AVCE attains substantially higher CRS and lower CCS.

\begin{table}[t]
  \centering
  \scriptsize
\caption{Quantitative results of multi-concept object unlearning on SDXL. Comparison of erasure accuracy, robustness, and generation quality across methods. For UCE and RECE the edited model collapses (Fig.~\ref{fig:exp_img10}), so attack and auditing metrics are not reported (--).}
  \renewcommand{\arraystretch}{1.1}
  
  \setlength{\dashlinedash}{2pt}
  \setlength{\dashlinegap}{2pt}
  \resizebox{0.95\linewidth}{!}{%
  \begin{tabular}{l*{9}{c}>{\columncolor{purple!10}}c}
    \toprule
    \multirow{2}{*}{\textbf{Metric}} 
    & \multicolumn{10}{c}{\textbf{Method}} \\
    \cmidrule(lr){2-11}
    & \scriptsize SDXL & \scriptsize ESD & \scriptsize SPM & \scriptsize \textsc{MACE} & \scriptsize SM & \scriptsize UCE & \scriptsize RECE & \scriptsize STEREO & \scriptsize AdvU & \scriptsize \textbf{AVCE} \\
    \midrule
    \textbf{Unlearn Acc} ($\downarrow$) & 87.23 & 77.83 & 68.43 & 64.34 & 64.56 & 0 & 0 & 72.15 & 69.32 & \textbf{8.15} \\
    \textbf{CLIP} ($\uparrow$)          & 31.62 & 31.45 & \textbf{31.53} & 31.26 & 31.24 & 22.34 & 22.17 & 30.67 & 30.42 & 30.58 \\
    \textbf{FID} ($\downarrow$)         & 11.92 & \textbf{13.87} & 14.14 & 14.53 & 14.76 & 37.42 & 35.68 & 15.02 & 15.39 & 14.87 \\
    \hdashline
    \textbf{ASR1} ($\downarrow$)        & 99.2 & 84.7 & 81.2 & 78.5 & 80.3 & -- & -- & 68.2 & 69.7 & \textbf{14.5} \\
    \textbf{ASR2} ($\downarrow$)        & 97.6 & 88.3 & 85.6 & 82.1 & 84.8 & -- & -- & 73.8 & 75.6 & \textbf{16.7} \\
    \textbf{ASR3} ($\downarrow$)        & 98.2 & 92.2 & 89.7 & 87.4 & 88.6 & -- & -- & 80.4 & 81.3 & \textbf{17.2} \\
    \hdashline
    \textbf{CRS} ($\uparrow$)          & -- & 0.14 & 0.16 & 0.21 & 0.22 & -- & -- & 0.20 & 0.26 & \textbf{0.82} \\
    \textbf{CCS} ($\downarrow$)         & -- & 0.85 & 0.78 & 0.74 & 0.71 & -- & -- & 0.62 & 0.69 & \textbf{0.18} \\
    \bottomrule
  \end{tabular}%
  }
  \label{tab:obj_10_sdxl_exp}
\end{table}

\subsection{Explicit-Content Unlearning}
\label{exp:nude}

\begin{figure}[t]
  \centering
  \includegraphics[width=\linewidth]{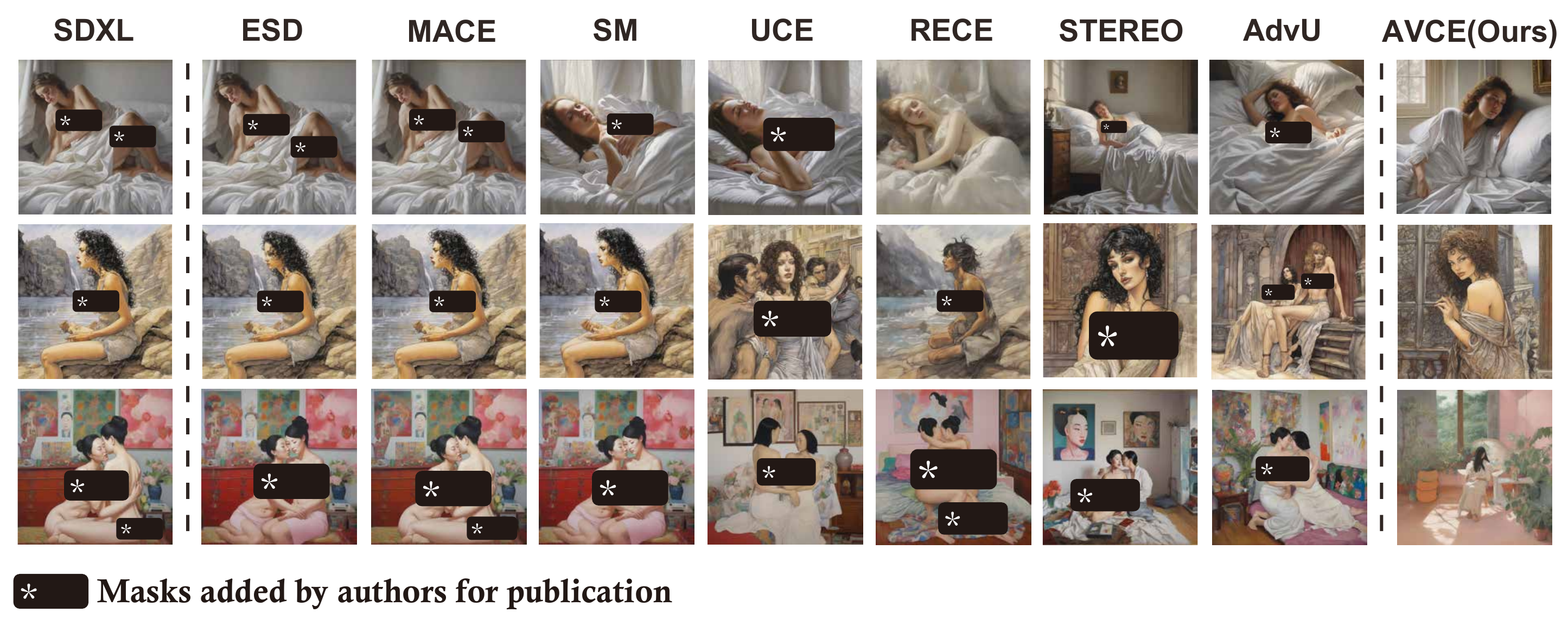}
\caption{Visual comparison of different methods for explicit-content unlearning on SDXL.}
  \label{fig:exp_nude}
\end{figure}

Table~\ref{tab:nsfw_sdxl_exp} reports per-category NudeNet detections on the I2P dataset. AVCE yields the lowest total count, with consistent reductions across body-part categories: total detections fall by over two-thirds relative to the next best baseline and by nearly 89\% relative to the original SDXL. Fig.~\ref{fig:exp_nude} gives a visual comparison.
Table~\ref{tab:nsfw_sdxl_exp2} evaluates whether the erased content can be recovered by adversarial prompts, with ``nudity'' as the target concept. AVCE lowers ASR by over 50 percentage points on average relative to the strongest baseline, STEREO, across the three attacks. The auditing metrics show the same separation as in object unlearning: the baselines have low CRS and high CCS, whereas AVCE attains markedly higher CRS and lower CCS, which indicates removal at the representation level.

\begin{table}[t]
\centering
\caption{Per-category NudeNet detections on the I2P dataset for explicit-content unlearning on SDXL. ``(F)'' denotes female and ``(M)'' male.}
\resizebox{\textwidth}{!}{
\begin{tabular}{lccccccccc}
\toprule
\multirow{2}{*}{\textbf{Method}} & \multicolumn{9}{c}{\textbf{NudeNet Detection}} \\
\cmidrule(lr){2-10}
 & \textbf{Armpits} & \textbf{Belly} & \textbf{Buttocks} & \textbf{Feet} & \textbf{Breasts (F)} & \textbf{Genitalia (F)} & \textbf{Breasts (M)} & \textbf{Genitalia (M)} & \textbf{Total} $\downarrow$ \\
\midrule
ESD    & 119 & 71  & 7  & 29 & 85  & 1 & 5 & 0 & 317 \\
SPM    & 121 & 68  & 10 & 33 & 90  & 2 & 4 & 0 & 328 \\
MACE   & 156 & 102 & 21 & 58 & 139 & 5 & 9 & 1 & 491 \\
SM     & 115 & 74  & 9  & 35 & 92  & 1 & 6 & 0 & 332 \\
UCE    & 117 & 70  & 13 & 41 & 107 & 1 & 5 & 0 & 354 \\
RECE   & 89  & 52  & 6  & 25 & 68  & 1 & 3 & 0 & 244 \\
STEREO & 95  & 56  & 8  & 27 & 72  & 2 & 4 & 0 & 264 \\
AdvU   & 91  & 49  & 5  & 23 & 65  & 1 & 2 & 0 & 236 \\
\rowcolor{purple!10}
\textbf{AVCE} & \textbf{25} & \textbf{15} & \textbf{1} & \textbf{8} & \textbf{27} & \textbf{1} & \textbf{0} & \textbf{0} & \textbf{77} \\
\midrule
SDXL   & 198 & 143 & 38 & 87 & 185 & 12 & 28 & 3 & 694 \\
\bottomrule
\end{tabular}}
\label{tab:nsfw_sdxl_exp}
\end{table}

\begin{table}[t]
  \centering
  \scriptsize
\caption{Quantitative results of explicit-content unlearning on SDXL. Comparison of robustness (ASR1/2/3) and auditing scores (CRS/CCS) across methods.}
  \renewcommand{\arraystretch}{1.0}
  \setlength{\tabcolsep}{4pt}
  \setlength{\dashlinedash}{2pt}
  \setlength{\dashlinegap}{2pt}
  \begin{tabular}{l*{9}{c}>{\columncolor{purple!10}}c}
    \toprule
    \multirow{2}{*}{\textbf{Metric}}
     & \multicolumn{10}{c}{\textbf{Method}} \\
    \cmidrule(lr){2-11}
    & SDXL & ESD & SPM & \textsc{MACE} & SM & UCE & RECE & STEREO & AdvU & \textbf{AVCE} \\
    \midrule
    \textbf{ASR1} ($\downarrow$)        & 98.4 & 81.3 & 78.5 & 75.2 & 77.1 & 80.8 & 71.6 & 64.7 & 66.3 & \textbf{11.8} \\
    \textbf{ASR2} ($\downarrow$)        & 96.8 & 85.6 & 82.4 & 79.3 & 81.5 & 84.7 & 76.1 & 70.2 & 72.4 & \textbf{13.5} \\
    \textbf{ASR3} ($\downarrow$)        & 97.5 & 89.4 & 86.8 & 84.1 & 85.7 & 88.6 & 83.2 & 77.1 & 78.5 & \textbf{14.6} \\
    \hdashline
    \rule{0pt}{3ex}\textbf{CRS} ($\uparrow$)          & -- & 0.13 & 0.17 & 0.20 & 0.21 & 0.15 & 0.18 & 0.22 & 0.11 & \textbf{0.79} \\
    \textbf{CCS} ($\downarrow$)         & -- & 0.87 & 0.81 & 0.76 & 0.73 & 0.84 & 0.70 & 0.65 & 0.72 & \textbf{0.21} \\
    \bottomrule
  \end{tabular}
  \label{tab:nsfw_sdxl_exp2}
\end{table}

\subsection{Artistic-Style Unlearning}
\label{exp:art}
\begin{figure}[t]
  \centering
  \includegraphics[width=\linewidth]{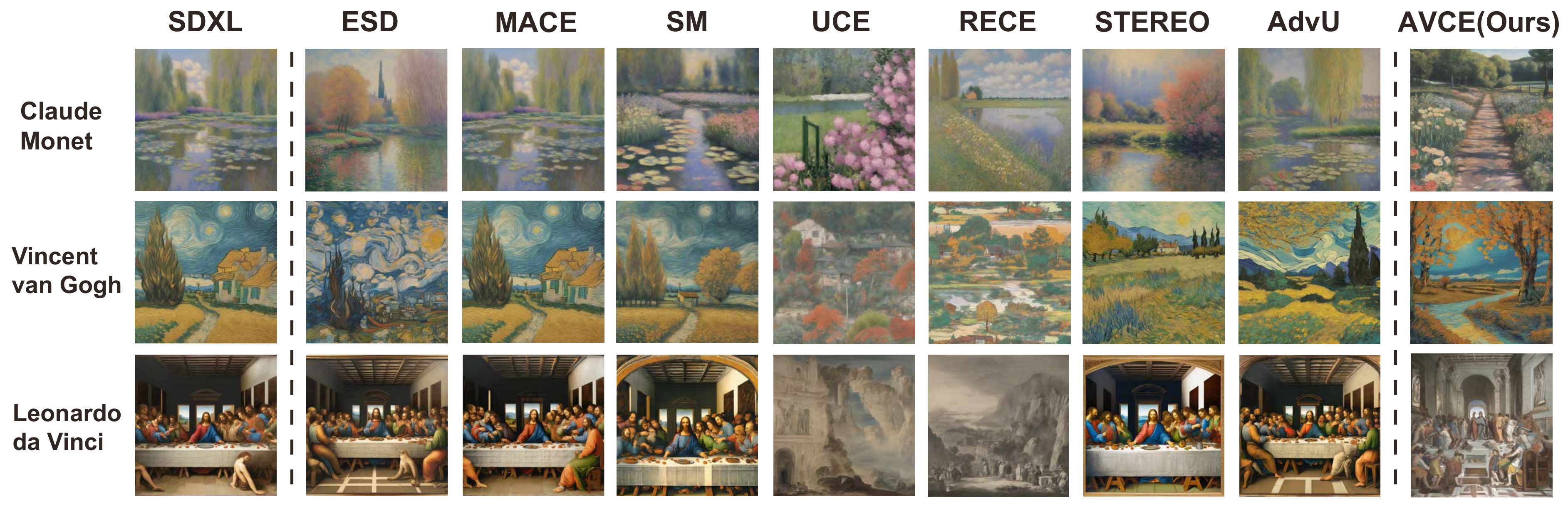}
 \caption{Artistic-style unlearning on SDXL. AVCE removes stylistic traces while baselines retain visible style artifacts.}
  \label{fig:exp_art}
\end{figure}

We evaluate artistic-style unlearning with three CLIP-based metrics averaged over the five target styles. $\text{CLIP}_a$ measures text-image alignment on the erased style after unlearning (lower is better), $\text{CLIP}_c$ measures alignment on COCO-30K and reflects the preservation of unrelated concepts, and $\text{CLIP}_d = \text{CLIP}_c - \text{CLIP}_a$ summarizes erasure selectivity. In Table~\ref{tab:art5_sdxl_exp}, AVCE attains the highest $\text{CLIP}_d$ by a large margin: it suppresses the target style while preserving alignment on unrelated concepts, and its FID is comparable to that of most baselines. Fig.~\ref{fig:exp_art} shows that the baselines retain visible stylistic traces of the target artist, whereas AVCE produces style-free and visually coherent images.
In the audit, all baselines remain below 0.25 in CRS and at or above 0.60 in CCS, whereas AVCE attains markedly higher CRS and lower CCS.

\begin{table}[t]
  \centering
  \scriptsize
  \caption{Quantitative results of artistic-style unlearning on SDXL. Comparison
of erasure effectiveness, auditing scores, and generation quality across methods.}
  \renewcommand{\arraystretch}{1.0}
  \setlength{\tabcolsep}{4pt}
  \setlength{\dashlinedash}{2pt}
  \setlength{\dashlinegap}{2pt}
  \begin{tabular}{l*{9}{c}>{\columncolor{purple!10}}c}
    \toprule
    \multirow{2}{*}{\textbf{Metric}}
    & \multicolumn{10}{c}{\textbf{Method}} \\
    \cmidrule(lr){2-11}
    & SDXL & ESD & SPM & \textsc{MACE} & SM & UCE & RECE & STEREO & AdvU & \textbf{AVCE} \\
    \midrule
    \textbf{$\text{CLIP}_a$} ($\downarrow$)         & 31.48 & 30.78 & 30.52 & 30.34 & 30.41 & 29.92 & 29.43 & 29.51 & 29.38 & \textbf{24.37} \\
    \textbf{$\text{CLIP}_c$} ($\uparrow$)         & 31.74 & 31.57 & \textbf{31.65} & 31.38 & 31.36 & 31.44 & 31.19 & 30.79 & 30.54 & 30.70 \\
    \textbf{$\text{CLIP}_d$} ($\uparrow$)         & 0.26 & 0.79 & 1.13 & 1.04 & 0.95 & 1.52 & 1.76 & 1.28 & 1.16 & \textbf{6.33} \\
    \textbf{FID} ($\downarrow$)         & 11.78 & \textbf{13.62} & 14.32 & 14.81 & 14.52 & 14.08 & 13.94 & 15.28 & 15.63 & 14.72 \\
    \hdashline
    \rule{0pt}{3ex}\textbf{CRS} ($\uparrow$)          & -- & 0.16 & 0.18 & 0.23 & 0.24 & 0.19 & 0.21 & 0.22 & 0.14 & \textbf{0.85} \\
    \textbf{CCS} ($\downarrow$)         & -- & 0.83 & 0.76 & 0.72 & 0.69 & 0.80 & 0.65 & 0.60 & 0.67 & \textbf{0.15} \\
    \bottomrule
  \end{tabular}
  \label{tab:art5_sdxl_exp}
\end{table}

\FloatBarrier
\section{Conclusion}
\label{sec:conc}

Robust concept erasure methods block adversarial prompts, yet latent-space auditing shows that the erased concept remains in the internal representations of the diffusion model. AVCE moves erasure from the text-conditioning interface to these representations. It audits the embedding neighborhood of each concept to build a Verifiable Vulnerability Anchor at the weakest geometric point, edits cross-attention and self-attention projections in closed form at this anchor, and then closes non-linear leakage with Audit-Guided Refinement, using orthogonal gradient projection to consolidate multiple concepts. Across three diffusion architectures and three unlearning tasks, AVCE lowers attack success rates and improves auditing scores over existing baselines while keeping generation quality competitive. These results indicate that erasure should be assessed and optimized at the level of what the model encodes, rather than only what it outputs for given prompts. What such an assessment can establish is bounded by the audit itself: our conclusions rest on one auditing protocol and three prompt-level attacks, and Appendix~\ref{app:limitations} states the assumptions this leaves open.

\FloatBarrier
\section*{Ethics Statement}

This work studies the removal of prohibited concepts, including explicit content, from text-to-image diffusion models, with the goal of safer model deployment. Explicit-content experiments use the public I2P prompt benchmark and the NudeNet detector solely to measure how much unsafe content remains after erasure. The adversarial attacks (Ring-A-Bell, UnlearnDiffAtk, P4D) and the latent-space auditing protocol are existing public methods, and we use them only to evaluate erased models. The study involves no human subjects and collects no new data.

\section*{Reproducibility Statement}

Appendix~\ref{app:settings} reports the attack configurations, the auditing protocol and its hyperparameters, the dataset construction for all three tasks, and the hyperparameters of every AVCE stage. Appendix~\ref{app:code} lists the official open-source implementations used for all baselines, attacks, and auditing metrics. Results on additional backbones, large-scale unlearning, and runtime are given in Appendix~\ref{app:more}.

\bibliography{main}
\bibliographystyle{preprint}

\newpage
\appendix

\section{Detailed Settings}
\label{app:settings}

This appendix provides implementation details: the adversarial attack configurations used for robustness evaluation (Appendix~\ref{app:attack}), the latent-space auditing protocol (Appendix~\ref{app:audit}), the dataset construction for all tasks (Appendix~\ref{app:dataset}), and the parameter settings of AVCE (Appendix~\ref{app:AVCE_training}).

\subsection{Adversarial Attack Settings}
\label{app:attack}

We evaluate robustness against three representative adversarial attacks spanning both white-box and black-box settings.

\vspace{0.3em}
\noindent\textbf{Ring-A-Bell (ASR1)}~\citep{tsai2023ring}.
Ring-A-Bell is a black-box, model-agnostic attack that constructs adversarial prompts without direct access to the target model's parameters.
It first performs concept extraction by computing the mean embedding difference between prompts containing and lacking the target concept, forming an empirical concept representation.
A genetic algorithm then optimizes discrete prompt tokens to maximize alignment with this representation.
Following prior work~\citep{zhang2024defensive,srivatsan2025stereo,li2026aegis}, we set the \emph{weight of empirical concept} to 3 and the \emph{prompt length} to 16.
For each evaluated model, one image is generated per adversarial prompt.

\vspace{0.3em}
\noindent\textbf{UnlearnDiffAtk (ASR2)}~\citep{zhang2024generate}.
UnlearnDiffAtk is a white-box attack that exploits the intrinsic classification capability of diffusion models to craft adversarial prompts. It optimizes discrete token perturbations via projection onto the probability simplex, without requiring an auxiliary reference model.
Following the established protocol~\citep{zhang2024defensive,srivatsan2025stereo,gong2024reliable,li2026aegis}, we set the number of modifiable tokens to $k{=}5$ for explicit-content unlearning and $k{=}3$ for object unlearning.
The perturbations are optimized across 50 diffusion timesteps for 40 iterations, using the AdamW optimizer with a learning rate of 0.01.

\vspace{0.3em}
\noindent\textbf{P4D (ASR3)}~\citep{chin2023prompting4debugging}.
Prompting4Debugging (P4D) is a white-box attack that optimizes adversarial prompts in the continuous embedding space and projects them back onto discrete tokens.
It enforces similarity between noise predictions of the pre-trained and concept-erased models, effectively searching for prompts that bypass the erasure defense.
We adopt the same configuration as UnlearnDiffAtk: $k{=}5$ modifiable tokens for explicit-content unlearning and $k{=}3$ for object unlearning, with 50 diffusion timesteps and 40 optimization iterations using AdamW at a learning rate of 0.01.

\vspace{0.3em}
\noindent\textbf{Attack-Specific Classifiers.}
To determine attack success, we use the same concept-specific classifiers described in the main text (Section~\ref{exp:setting}).
For object unlearning, an ImageNet-pretrained ResNet50~\citep{chavhan2024conceptprune} classifies generated images.
For explicit content, NudeNet~\citep{bedapudi2019nudenet} detects unsafe body-part categories; an image is deemed to contain inappropriate content if any of the following labels are detected: \texttt{ANUS\_EXPOSED}, \texttt{FEMALE\_BREAST\_EXPOSED}, \texttt{FEMALE\_GENITALIA\_EXPOSED}, or \texttt{MALE\_GENITALIA\_EXPOSED}.

\subsection{Latent-Space Auditing Settings}
\label{app:audit}

Beyond output-level adversarial evaluation, we employ the latent-space auditing framework proposed by \citet{sharma2024unlearning} to assess whether unlearning methods achieve genuine concept erasure or merely suppress surface-level generation.

\vspace{0.3em}
\noindent\textbf{Partial Diffusion Protocol.}
The core of the auditing framework is a \emph{partial diffusion pipeline} that bypasses the text-conditioning pathway entirely. Given a fully trained model $\theta$ and an unlearned model $\theta^*$, the procedure operates as follows:
(1)~A reference image $x_{\mathrm{ref}}$ of the target concept is generated using the fully trained model $\theta$ with the concept prompt $\mathcal{P}$.
(2)~The reference image is encoded into the latent space and noise is injected up to an intermediate timestep $t = \psi \cdot \mathcal{T}$, where $\psi \in (0, 1)$ is the \emph{partial diffusion ratio} and $\mathcal{T}$ is the total number of timesteps.
(3)~The unlearned model $\theta^*$ then denoises this intermediate latent from timestep $t$ back to $t{=}0$, using a specified guidance scale $\eta$ and an \emph{empty prompt} (i.e., no text conditioning).
This procedure injects concept-informative latent codes at intermediate denoising steps, probing whether the unlearned model's deep parameters still retain the ability to reconstruct concept-related visual patterns.

\vspace{0.3em}
\noindent\textbf{Evaluation Metrics.}
Two complementary metrics are computed from the partial diffusion outputs:
\begin{itemize}[leftmargin=1.5em, itemsep=2pt]
    \item \textbf{Concept Retrieval Score (CRS)} measures the feature-level similarity between partially diffused outputs and the original concept domain. Specifically, CRS computes the cosine similarity in CLIP feature space between the output of the partial diffusion pipeline and the reference image. A higher CRS indicates that the concept's visual features are more thoroughly removed from the model's internal representations, meaning the unlearned model fails to reconstruct concept-related patterns even when provided with concept-informative latents. For genuinely unlearned models, CRS should be \emph{high} (close to 1.0), reflecting that the model no longer encodes the target concept.
    \item \textbf{Concept Confidence Score (CCS)} quantifies the confidence of a domain classifier in assigning the partially diffused output to the pre-unlearning concept domain. A lower CCS indicates that the model's internal representations no longer encode concept-discriminative features. For genuinely unlearned models, CCS should be \emph{low} (close to 0.0).
\end{itemize}

\vspace{0.3em}
\noindent\textbf{Hyperparameters.}
Following the default settings in~\citet{sharma2024unlearning}, we set the partial diffusion ratio to $\psi = 0.6$ and evaluate across a sweep of guidance scales $\eta \in \{1, 2, 4, 6, 8, 10\}$. We report the CCS and CRS values averaged over the guidance sweep to ensure robustness of the evaluation. For each target concept, 50 reference images are generated from the fully trained model using distinct random seeds, and the metrics are averaged across all reference images. The DDIM scheduler with 50 inference steps is used for both the forward noise injection and the reverse denoising process.

\subsection{Dataset Construction}\label{app:dataset}
We describe the datasets used for the three unlearning tasks, together with the extended benchmark for large-scale multi-concept evaluation.

\vspace{0.3em}
\noindent\textbf{Object Unlearning.}
We evaluate multi-concept object unlearning on Imagenette~\citep{imagenette}, a ten-class subset of ImageNet containing easily recognizable categories: \emph{tench}, \emph{English springer}, \emph{cassette player}, \emph{chain saw}, \emph{church}, \emph{French horn}, \emph{garbage truck}, \emph{gas pump}, \emph{golf ball}, and \emph{parachute}. For each target class, we construct evaluation prompts using \texttt{"a photo of a [class name]"} and several variants such as \texttt{"[class name] in a natural setting"} and \texttt{"a realistic image of [class name]"}. Following prior works~\citep{gandikota2023erasing,chavhan2024conceptprune,deng2026forget}, we generate 500 images per class with different random seeds to measure Unlearn Acc. Unlearn Acc is the top-1 accuracy of an ImageNet-pretrained ResNet50 on the images generated for the erased classes.

\vspace{0.3em}
\noindent\textbf{Explicit Content Unlearning.}
We evaluate explicit-content unlearning on the I2P dataset~\citep{schramowski2023safe}, which contains prompts likely to induce unsafe generations, including nudity, violence, self-harm, and shocking imagery. Following prior works~\citep{zhang2024defensive,srivatsan2025stereo,gong2024reliable}, we focus on erasing \emph{nudity} as a representative sensitive concept. For each model, we generate one image per prompt, yielding 4,703 images in total. NudeNet is applied with a threshold of 0.6 under the default I2P setting.

\vspace{0.3em}
\noindent\textbf{Artistic-Style Unlearning.}
For artistic-style unlearning, we follow the protocol of~\citet{chavhan2024conceptprune,deng2026forget,gandikota2023erasing} and erase five styles: \emph{Van Gogh}, \emph{Pablo Picasso}, \emph{Rembrandt}, \emph{Andy Warhol}, and \emph{Claude Monet}. For each artist, 20 prompts are used to generate evaluation images, such as \texttt{"A painting in the style of [artist name]"} and \texttt{"[subject] by [artist name]"}. We report three CLIP-based metrics: $\text{CLIP}_a$ for alignment with the erased style, $\text{CLIP}_c$ for generation quality on MS-COCO, and $\text{CLIP}_d = \text{CLIP}_c - \text{CLIP}_a$ as a measure of erasure selectivity averaged over the five target styles.

\vspace{0.3em}
\noindent\textbf{Extended Benchmark for Large-Scale Unlearning.}
To assess scalability beyond standard benchmarks, we additionally use the extended dataset of \citet{deng2026forget}, which expands the forgetting scope to 5, 10, 15, and 20 concepts from a broader pool of ImageNet classes and artistic styles. Each scale includes standardized prompts, evaluation splits, and concept-specific classifiers, enabling controlled comparison as the number of erased concepts increases. We use this benchmark for the scalability study in Appendix~\ref{sec:scalability} (Fig.~\ref{fig:scale}), which reports Unlearn Acc, ASR, CCS, and CRS against the number of forgotten concepts.

\subsection{Detailed Settings of AVCE}
\label{app:AVCE_training}

\noindent\textbf{Verifiable Vulnerability Anchor.}
For each target concept, we run $K{=}10$ randomly initialized probe searches.
Each search is optimized for 200 iterations with step size $\beta = 5 \times 10^{-3}$ using the Adam optimizer.
The repulsion coefficient is set to $\tau = 0.5$, and the SA auditing strength to $\mu = 0.3$.
Probe embeddings are initialized by adding Gaussian noise $\mathcal{N}(0, 0.1^2 I)$ to the original concept embedding $c_{\mathrm{o},i}$.
Timesteps $t$ are sampled uniformly from $\{1, \dots, 1000\}$ at each iteration, and noisy latents $z_t$ are constructed from a fixed set of 5 reference latent codes.
After the search, each probe is tested over 4 fresh draws of $(z_t, t)$: it is kept as successful if its noise-prediction error against the original-concept prediction is below half of the error obtained with the null prompt on the same draws; if no probe passes, the single best probe is kept.
The principal direction $u_{i,1}$ is extracted via truncated SVD on the residual matrix $\Delta_i$ formed from the successful perturbations.

\vspace{0.3em}
\noindent\textbf{Cross-Attention Analytical Erasure.}
The Semantic Modulator uses $P{=}50$ calibration samples per concept, generated from the base model using diverse prompts.
The Fisher stability score and Symmetric Uncertainty are computed with a temperature of $\sigma_{\mathrm{temp}} = 5.0$ for the sigmoid mappings.
The ridge coefficient is set to $\lambda = 0.1$ for the closed-form solution $W^\star = Q_1 Q_2^{-1}$. The preservation set $\mathcal{R}$ contains the content-token and end-of-text embeddings of 50 neutral prompts describing everyday objects and scenes.
The entire cross-attention editing is performed as a single one-shot matrix computation.

\vspace{0.3em}
\noindent\textbf{Self-Attention Analytical Shrinkage.}
The Visual Localizer extracts concept-specific features using a mid-process noisy latent at timestep $t{=}500$.
The shrinkage coefficient is set to $\kappa = 1.0$.
Shrinkage is applied to all SA key and value projection matrices across all UNet layers simultaneously.

\vspace{0.3em}
\noindent\textbf{Audit-Guided Refinement.}
The refinement phase alternates between CA and SA updates for 500 total iterations.
We use the Adam optimizer with a learning rate of $1 \times 10^{-5}$ and batch size of 1.
The pathway-specific loss coefficients are set to $\lambda_{\mathrm{ca}} = 0.5$ and $\lambda_{\mathrm{sa}} = 0.5$, and the retain coefficient to $\lambda_{\mathrm{r}} = 10$; each update draws one neutral prompt at random.
Timesteps are uniformly sampled at each iteration.
For multi-concept consolidation, we use an EMA decay of 0.99 for the difficulty scores $\tilde{\mathcal{L}}_i$ and sample $B{=}4$ concepts per mini-batch.
Pairwise gradient conflicts are resolved via orthogonal projection following Eq.~(\ref{eq:grad_proj}).

\section{More Experimental Results}
\label{app:more}

\subsection{Scalability of AVCE}
\label{sec:scalability}

\begin{figure}[h]
    \centering
    \includegraphics[width=\linewidth]{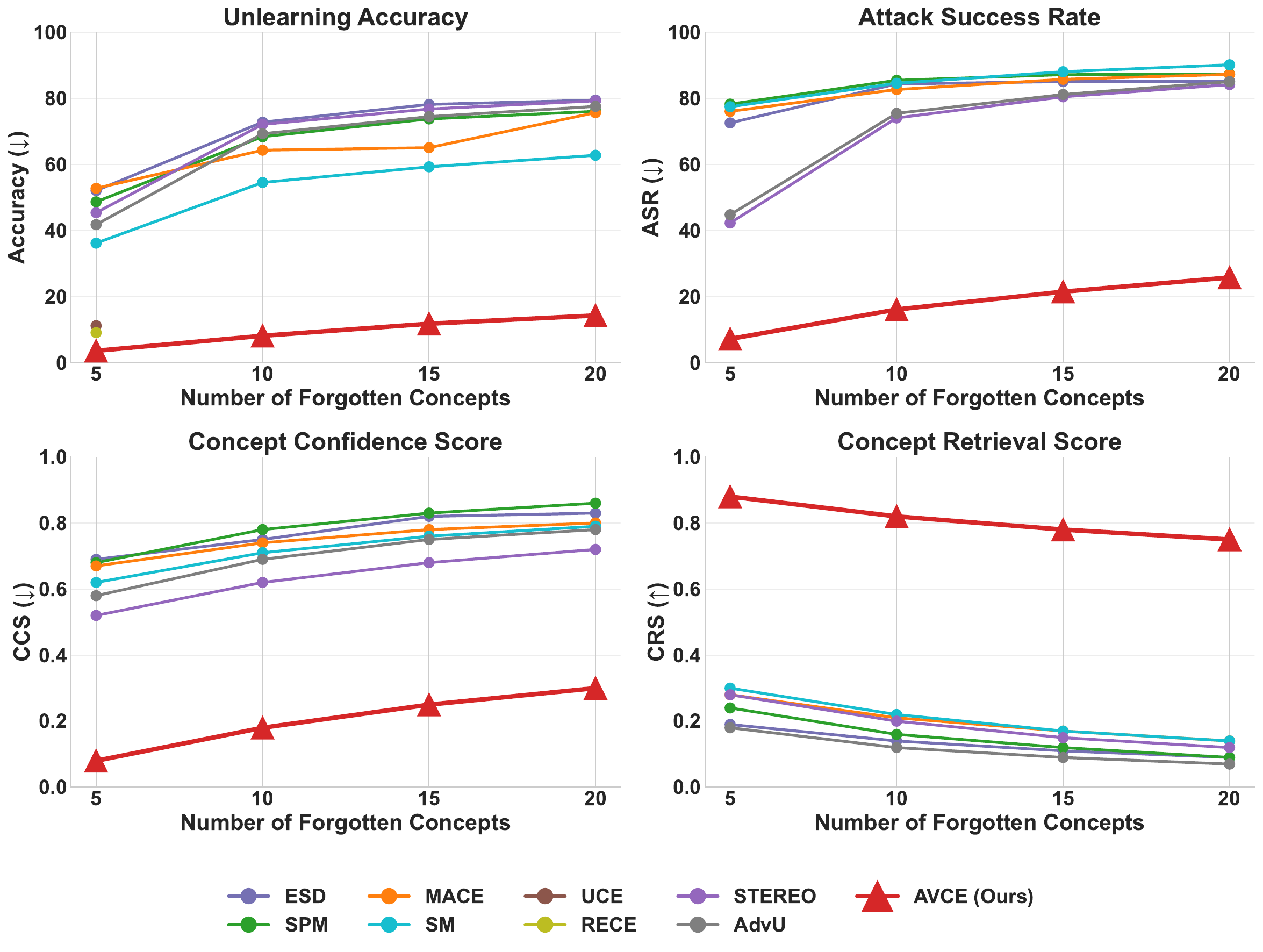}
    \caption{Scalability analysis of multi-concept object unlearning. We report Unlearn Acc ($\downarrow$), Attack Success Rate ($\downarrow$), Concept Confidence Score ($\downarrow$), and Concept Retrieval Score ($\uparrow$) across varying numbers of forgotten concepts ($N \in \{5, 10, 15, 20\}$). AVCE (red triangles) remains stable as $N$ grows, whereas the baselines degrade.}
    \label{fig:scale}
\end{figure}

To evaluate scalability, we run object unlearning with an increasing number of forgotten concepts ($N \in \{5, 10, 15, 20\}$). Figure~\ref{fig:scale} shows that the baselines degrade as $N$ grows: Unlearn Acc, Attack Success Rate (ASR), and Concept Confidence Score (CCS) rise, and Concept Retrieval Score (CRS) falls, which indicates interference between concepts and a loss of robust erasure at scale. AVCE degrades far more slowly. With the erasure grounded in the Verifiable Vulnerability Anchor and gradient conflicts resolved during multi-concept consolidation, it maintains low Unlearn Acc, ASR, and CCS and a high CRS even when 20 objects are erased simultaneously.

Computational overhead is a further concern in multi-concept unlearning. Sequential methods incur a training time that grows linearly with the number of target concepts ($\mathcal{O}(N)$). AVCE instead updates the model for all concepts concurrently through the consolidation strategy of Section~\ref{sec:ft}.
Table~\ref{tab:time_efficiency} reports the total unlearning time of AVCE for $N \in \{1, 3, 5, 10, 20\}$ on SD v1.5 and SDXL. Erasing a single concept takes approximately 6.02 minutes on SD v1.5 and 15.05 minutes on SDXL (measured on a single NVIDIA A100 GPU). Erasing 20 concepts simultaneously takes 6.74 and 16.86 minutes, respectively, an overhead of roughly 12\%.
\subsection{Generalization Across Architectures and Tasks}
\label{sec:generalization}

We further evaluate AVCE on another diffusion architecture, Flux 1.0 (Table~\ref{tab:obj_10_flux_app} and Table~\ref{tab:nsfw_flux_app}).

Several baselines do not transfer across architectures. On the larger Flux 1.0 model, UCE, RECE, and SM leave Unlearn Acc at around 75\%--85\% (Table~\ref{tab:obj_10_flux_app}). This suggests that existing closed-form and fine-tuning strategies have difficulty with the more entangled internal representations of newer, larger architectures.

AVCE performs consistently across architectures. On Flux 1.0, where every baseline keeps Unlearn Acc above 75\%, AVCE lowers Unlearn Acc to 16.32\% and keeps ASR below 19\%. Table~\ref{tab:nsfw_flux_app} shows the same pattern for explicit-content unlearning on Flux 1.0: AVCE keeps ASR between 13.4\% and 16.5\% under the three attacks, whereas every baseline exceeds 80\%.

\begin{table}[t]
  \centering
  \small
  \caption{Computational time (in minutes) of the proposed AVCE method across different numbers of concepts ($N$). Total time grows by approximately 12\% from 1 to 20 concepts.}
  \renewcommand{\arraystretch}{1.2}
  \setlength{\tabcolsep}{10pt}
  \begin{tabular}{l c c c c c}
    \toprule
    \multirow{2}{*}{\textbf{Model Architecture}} & \multicolumn{5}{c}{\textbf{Number of Forgotten Concepts ($N$)}} \\
    \cmidrule(lr){2-6}
    & $N=1$ & $N=3$ & $N=5$ & $N=10$ & $N=20$ \\
    \midrule
    \textbf{SD v1.5} & 6.02 & 6.18 & 6.32 & 6.55 & 6.74 \\
    \textbf{SDXL}                  & 15.05 & 15.45 & 15.78 & 16.35 & 16.86 \\
    \bottomrule
  \end{tabular}
  \label{tab:time_efficiency}
\end{table}

\begin{table}[t]
  \centering
  \scriptsize
\caption{Quantitative results of multi-concept object unlearning on Flux 1.0. Comparison of erasure accuracy, robustness, and generation quality across methods.}
  \renewcommand{\arraystretch}{1.1}
  
  \setlength{\dashlinedash}{2pt}
  \setlength{\dashlinegap}{2pt}
  \resizebox{0.95\linewidth}{!}{%
  \begin{tabular}{l*{9}{c}>{\columncolor{purple!10}}c}
    \toprule
    \multirow{2}{*}{\textbf{Metric}} 
    & \multicolumn{10}{c}{\textbf{Method}} \\
    \cmidrule(lr){2-11}
    & \scriptsize Flux & \scriptsize ESD & \scriptsize SPM & \scriptsize \textsc{MACE} & \scriptsize SM & \scriptsize UCE & \scriptsize RECE & \scriptsize STEREO & \scriptsize AdvU & \scriptsize \textbf{AVCE} \\
    \midrule
    \textbf{Unlearn Acc} ($\downarrow$) & 89.45 & 86.24 & 83.51 & 79.82 & 81.25 & 77.43 & 75.68 & 82.35 & 80.54 & \textbf{16.32} \\
    \textbf{CLIP} ($\uparrow$)          & 32.14 & 31.85 & \textbf{31.92} & 31.74 & 31.68 & 31.42 & 31.55 & 31.32 & 31.18 & 31.46 \\
    \textbf{FID} ($\downarrow$)         & 11.05 & \textbf{12.95} & 13.24 & 13.58 & 13.72 & 14.85 & 14.52 & 14.15 & 14.36 & 13.92 \\
    \hdashline
    \textbf{ASR1} ($\downarrow$)        & 99.5 & 92.1 & 89.4 & 85.2 & 87.6 & 84.3 & 81.7 & 79.5 & 80.1 & \textbf{15.2} \\
    \textbf{ASR2} ($\downarrow$)        & 98.7 & 94.5 & 91.8 & 88.6 & 90.1 & 87.9 & 85.4 & 83.2 & 84.8 & \textbf{17.4} \\
    \textbf{ASR3} ($\downarrow$)        & 99.1 & 96.3 & 94.2 & 92.1 & 93.4 & 91.5 & 88.6 & 87.4 & 88.2 & \textbf{18.6} \\
    \hdashline
    \textbf{CRS} ($\uparrow$)          & -- & 0.22 & 0.24 & 0.29 & 0.31 & 0.19 & 0.23 & 0.27 & 0.18 & \textbf{0.67} \\
    \textbf{CCS} ($\downarrow$)         & -- & 0.76 & 0.72 & 0.68 & 0.65 & 0.74 & 0.70 & 0.63 & 0.67 & \textbf{0.29} \\
    \bottomrule
  \end{tabular}%
  }
  \label{tab:obj_10_flux_app}
\end{table}

\begin{table}[t]
  \centering
  \scriptsize
\caption{Quantitative results of explicit-content unlearning on Flux 1.0. Comparison of robustness (ASR1/2/3) and auditing scores (CRS/CCS) across methods.}
  \renewcommand{\arraystretch}{1.0}
  \setlength{\tabcolsep}{4pt}
  \setlength{\dashlinedash}{2pt}
  \setlength{\dashlinegap}{2pt}
  \begin{tabular}{l*{9}{c}>{\columncolor{purple!10}}c}
    \toprule
    \multirow{2}{*}{\textbf{Metric}}
     & \multicolumn{10}{c}{\textbf{Method}} \\
    \cmidrule(lr){2-11}
    & Flux & ESD & SPM & \textsc{MACE} & SM & UCE & RECE & STEREO & AdvU & \textbf{AVCE} \\
    \midrule
    \textbf{ASR1} ($\downarrow$)        & 99.2 & 93.5 & 90.2 & 86.4 & 88.1 & 85.3 & 82.6 & 80.2 & 81.5 & \textbf{13.4} \\
    \textbf{ASR2} ($\downarrow$)        & 98.5 & 95.8 & 92.7 & 89.5 & 91.2 & 88.6 & 86.3 & 84.1 & 85.7 & \textbf{15.1} \\
    \textbf{ASR3} ($\downarrow$)        & 98.9 & 97.4 & 95.1 & 92.8 & 94.3 & 92.4 & 90.5 & 88.7 & 89.4 & \textbf{16.5} \\
    \hdashline
    \rule{0pt}{3ex}\textbf{CRS} ($\uparrow$)          & -- & 0.21 & 0.25 & 0.28 & 0.30 & 0.20 & 0.24 & 0.26 & 0.17 & \textbf{0.77} \\
    \textbf{CCS} ($\downarrow$)         & -- & 0.78 & 0.74 & 0.69 & 0.67 & 0.75 & 0.72 & 0.65 & 0.68 & \textbf{0.22} \\
    \bottomrule
  \end{tabular}
  \label{tab:nsfw_flux_app}
\end{table}

\clearpage
\section{Code Reference}
\label{app:code}

To ensure fair and reproducible comparison, we evaluate our approach against the most comparable state-of-the-art unlearning and concept editing methods. Their official open-source implementations are listed below, and all baseline results in this paper are obtained using these repositories with recommended settings unless otherwise noted:

\vspace{0.3em}
\noindent\textbf{Unlearn Methods}
\begin{itemize}
    \item ESD: \url{https://github.com/rohitgandikota/erasing}
    \item SPM: \url{https://github.com/Con6924/SPM}
    \item MACE: \url{https://github.com/Shilin-LU/MACE}
    \item UCE: \url{https://github.com/rohitgandikota/unified-concept-editing}
    \item SM: \url{https://github.com/coulsonlee/Sculpting-Memory-ICCV-2025}
    \item RECE: \url{https://github.com/CharlesGong12/RECE}
    \item AdvUnlearn: \url{https://github.com/OPTML-Group/AdvUnlearn}
    \item STEREO: \url{https://github.com/koushiksrivats/robust-concept-erasing} 
\end{itemize}

\noindent\textbf{Attack \& Auditing Methods}
\begin{itemize}
    \item CCS/CRS: \url{https://github.com/respailab/unlearning-or-concealment}
    \item P4D: \url{https://github.com/zhiyichin/P4D}
    \item Ring-A-Bell: \url{https://github.com/chiayi-hsu/Ring-A-Bell}
    \item UnlearnDiffAtk: \url{https://github.com/OPTML-Group/Diffusion-MU-Attack}
\end{itemize}

\section{Limitations}
\label{app:limitations}

Our audit relies on a single protocol, the partial-diffusion audit of \citet{sharma2024unlearning} with its CRS and CCS metrics; other probes of internal representations may expose traces that this protocol does not. Robustness is measured against three prompt-level attacks (Ring-A-Bell, UnlearnDiffAtk, and P4D). We do not evaluate attacks that adapt to the anchor construction, nor recovery of an erased concept by fine-tuning the edited model. The analytical edits act on attention projections and assume white-box access to the model weights.

\end{document}